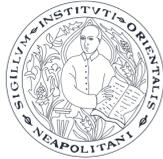

Università degli studi di Napoli
"L'Orientale"

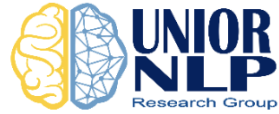

# *#LaCulturaNonSiFerma*
# Report on Use and Diffusion of #Hashtags from the Italian Cultural Institutions during the COVID-19 outbreak

**Authors**: Carola Carlino, Gennaro Nolano, Maria Pia di Buono, Johanna Monti - University of Naples "L'Orientale" - UNIOR NLP Research Group

**Date**: 22/03/2021
**Version**: 1.5

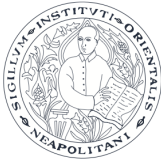

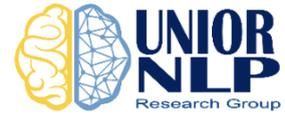

*Note: This report is the EN reduced version of the IT report submitted to arXiv on the 21st of May 2020 and available at* [*https://arxiv.org/abs/2005.10527*](https://arxiv.org/abs/2005.10527)

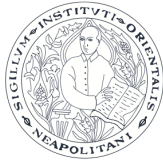

Università degli studi di Napoli
"L'Orientale"

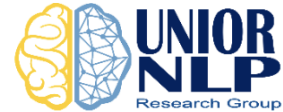

# Introduction

During the COVID-19 outbreak, spread between the end of 2019 and the beginning of 2020, many countries applied a lockdown policy, still ongoing in some places if the time frame in which the report has been compiled is considered. These restrictive measures interested billions of citizens, institutions, and companies that had to suspend their activity in order to contain and contrast the virus spreading. Italy was one of the first world nations to adopt these measures that involved cultural institutions as well.

Museums and art galleries adopted several strategies to continue engaging their audience, and, in most cases, they chose to increase their communication on the main social platforms, like Facebook, Twitter and Instagram. An active users' participation was requested to reach two main goals: 1) keeping the communication channel active for the audience 2) let the people who had to stay at home enjoy cultural contents.

This report presents a first analysis of the *#hashtags* that Italian cultural institutions produced or used during the time period from the 8th of March 2020 to the 4th of May 2020. The selected time span refers to the Italian lockdown period, as the first date corresponds to the publication of the administrative order 'DPCM March 8th 2020' stating the so-called 'Phase 1' to start, while the second one corresponds to the publication of the administrative order 'DPCM April 26th 2020' with the announcement to restriction loosening during the 'Phase 2'.

The choice of focusing on the analysis of *#hashtags* arises from their characteristic of thematic aggregators which contribute to create archives of information related to a specific topic, stored in a digital environment.

# Data Collection and Analysis

This analysis takes into account the *#hashtags* that Italian cultural institutions used on Twitter to communicate their initiatives and their proposals to the audience during Phase 1. According to a survey conducted in 2016 by the Digital Innovation Observatory for Cultural Heritage, the percentage of Italian museums using Twitter was about 31%[1]. The use of this platform increased to 33%[2] over the following two years (2017-18). The report compiled by the Network European Museum Organisations (NEMO) observed that, after three weeks the beginning of the pandemic, the 80% of Italian museums increased the online activity, as an attempt to enhance cultural content fruition[3].

In order to collect the main *#hashtags* of Twitter posts published by the Italian museums and by the Italian Ministry for Cultural Heritage and Activity and for Tourism (MiBACT), we have

---

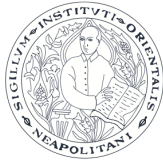

manually selected a set of relevant topics, among the ones proposed as 'trend topics' on Twitter. Then a first dataset has been created including all the tweets published in the period of reference together with one or more of the selected #hashtags (Table 1). This dataset contains 23716 tweets, collected by using Python 3.6 and the GetOldTweets library[4].

Subsequently, the #hashtags in the dataset have been grouped in two subgroups: 1) those existing before the COVID-19 diffusion and the introduction of the administrative decrees and 2) those created during the lockdown period.

Then, in order to reduce the dataset to a more focused selection of tweets, we filter out the #hashtags presenting less than 1000 tweets and the ones not directly related to the Cultural Heritage domain (e.g., #quarantinelife). For this reason, six #hashtags have been chosen from the total amount; some of them, e.g., #museitaliani, existed before the lockdown period, but they were re-used by museums in the period of reference. The final dataset is composed of 15988 tweets containing at least one of the selected #hashtags.

In Table 2 a summary of the interactions per #hashtag is reported. The interaction corresponds to the sum of tweets and retweets for each #hashtag. It can be observed that #ArTyouReady is the most popular #hashtag, since it counts 16237 retweets during Phase 1. This #hashtag was created during the pandemic by MiBACT to launch a campaign aimed at bringing on social networks places of culture temporarily closed through pictures posted by users. The initiative ran from the 29th of March to the 26th of April, renewed every week by different requests from time to time.

Finally, we collect some information about users by means of Tweepy library[5], namely verified users and their geographical location. Among the verified users, we identify the accounts owned by Italian cultural institutions and organisations.

Table 1 - Retweets per# hashtag. The #hashtags in grey are those created before the COVID-19 pandemic.

| #Hashtag | Retweet |
|---|---|
| #artyouready | 20142 |
| #artchallenge | 16282 |
| #emptymuseum | 12966 |
| #cultureinquaratine | 11888 |
| #museitaliani | 10505 |
| #museichiusimuseiaperti | 6470 |
| #laculturanonsiferma | 3502 |
| #laculturaincasa | 2566 |
| #shareourheritage | 2305 |
| #resiliart | 1845 |
| #museumfromhome | 1833 |
| #lartetisomiglia | 1540 |
| #museumalphabet | 1469 |
| #museumfromhome | 1183 |
| #museumgames | 1171 |
| #laculturaincasaKIDS | 599 |
| #laculturacura | 540 |
| #ilmuseoacasa | 379 |
| #ilmuseoacasatua | 316 |
| #museumathome | 202 |
| #museumquiz | 191 |
| #ITweetMuseums | 180 |
| #digitalflashmob | 115 |
| #lartenonsiferma | 115 |
| #nomicoseneimusei | 111 |
| #quarantinelife | 97 |
| #betweenartandquarantine | 89 |
| #ABCBarberiniCorsini | 44 |
| #museiitaliani | 26 |
| #condividilacultura | 21 |
| #stayathomechallenge | 14 |
| #tussenkunstenquarantine | 10 |
| #iogiocodacasa | 6 |

Table 2 - Interaction per #hashtag

| Promoter Institution | Release Date | #Hashtag | Tweet | Retweet | Total |
|---|---|---|---:|---:|---:|
| MiBACT | 29/03 | #artyouready | 5310 | 20142 | 25452 |
| MiBACT | 29/03 | #emptymuseum | 3204 | 12966 | 16170 |
| Direzione generale Musei - MiBACT | Before 2020 | #museitaliani | 1874 | 10505 | 12379 |
| Museo Tattile di Varese | 24/02 | #museichiusimuseiaperti | 2639 | 6470 | 9109 |
| ICOM  Italia | 07/03 | #laculturanonsiferma | 1732 | 3502 | 5234 |
| Musei In Comune (MIC) | 17/03 | #laculturaincasa | 1229 | 2566 | 3795 |
| **Totale** | | | **15988** | **56151** | **72139** |

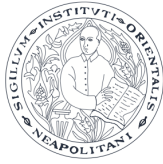

Università degli studi di Napoli
"L'Orientale"

Diagram in Figure 1 reports the number of tweets for each *#hashtag* we consider, spanning the reference period. It is worth stressing that on the 29th March, the diagram shows a peak for the *#hashtag #artyouready*, followed by *#emptymuseum*, while a more or less moderate use of all the *#hashtags* has been recorded.

This peak can be explained by the launch of the *#ArTyouReady* campaign.

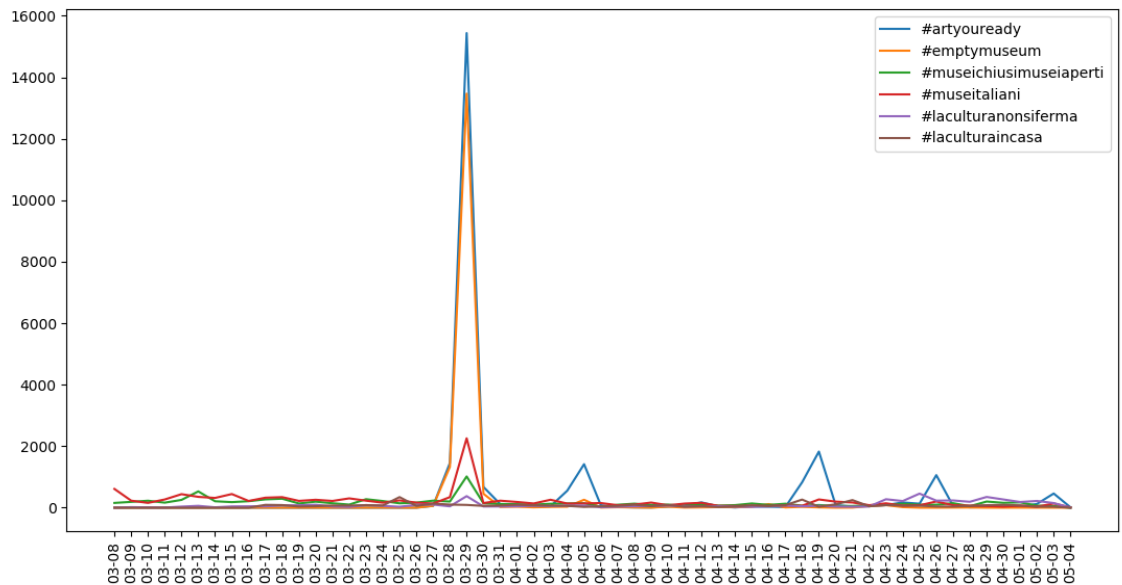

Figure 1 - Interaction trend per #hashtag

**Spread of #hashtags.** Figure 2 reports the *#hashtags*' trend used in March. Following a short increase observed on the 13th of March, the flow is flat until the 28th of March, when the line starts growing to reach the peak on that day, due to the *#ArTyouReady* initiative.

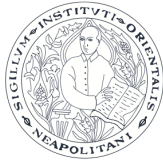

Università degli studi di Napoli
"L'Orientale"

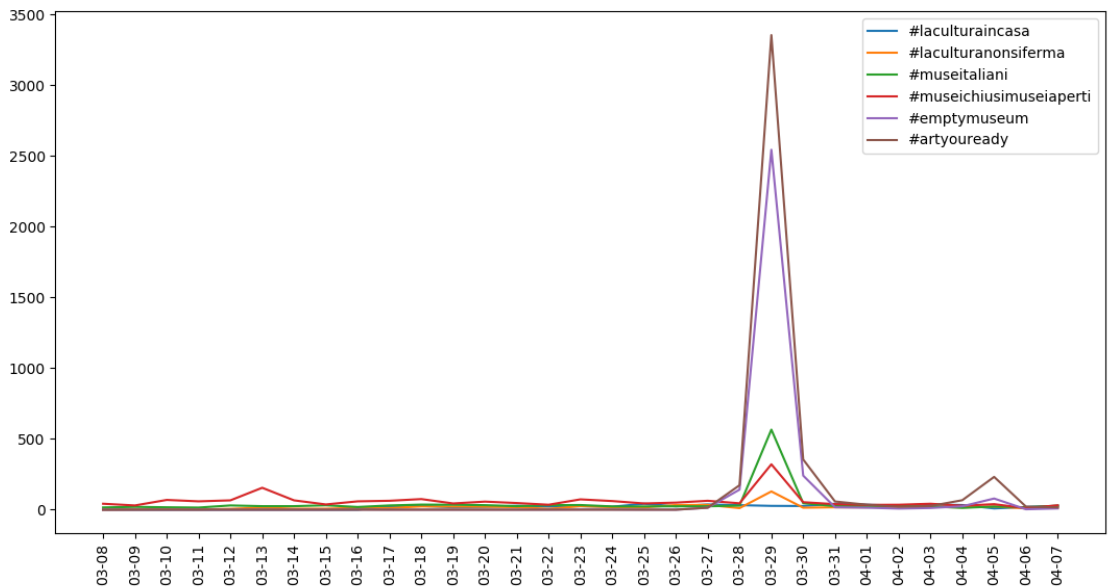

Figure 2 - #hashtags trend among the tweets created during the month of March

Figure 3 reports the *#hashtags*' trend related to the tweets produced in April. Among the analysed ones, an unvaried trend can be observed for *#emptymuseum*, with a short variation on the 23rd April.

A stable increase is revealed for *#museichiusimuseiaperti,* although a considerable variation has been registered from the 29th of April to the 1st of May. On the 19th and 26th of April, and the 3rd of May, we notice an increase of *#ArTyouReady*, due to the fact that the initiative related to this *#hashtag* has been relaunched by MiBACT, as proved by the same trend reported for the days immediately preceding.

For all the other labels, there is a considerable increase between the end of April and the beginning of May, namely from the 22nd of April to the 4th of May.

Figure 4 shows the retweets' trend per *#hashtag* produced in March. An increase in activities can be noted in correspondence of the 28th - 30th of March, and a slight increase with reference to the beginning of April.

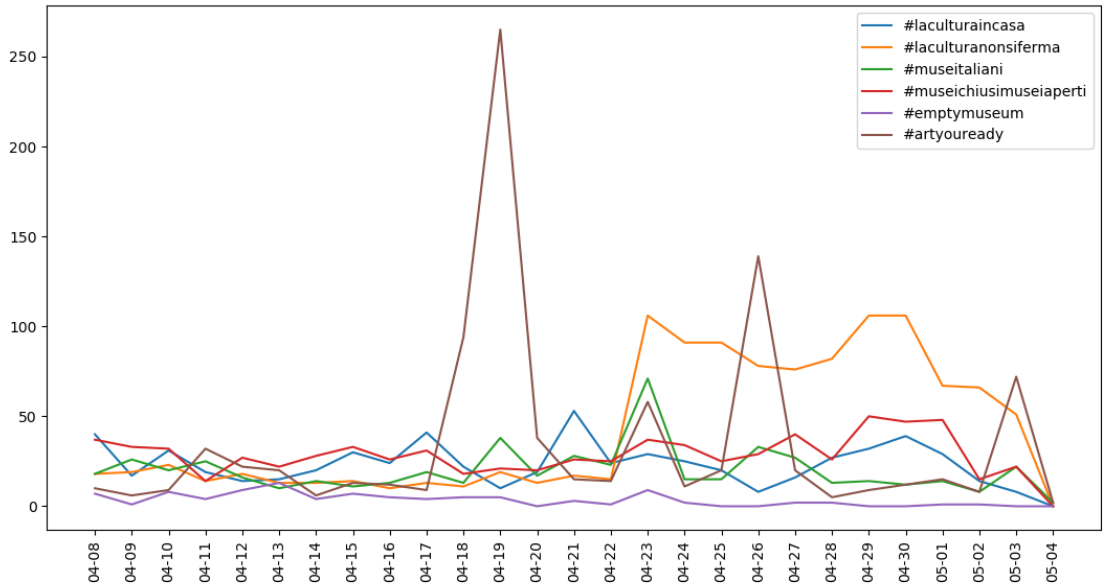

Figure 3 - #Hashtags' trend among the tweets created during the month of April and the early May

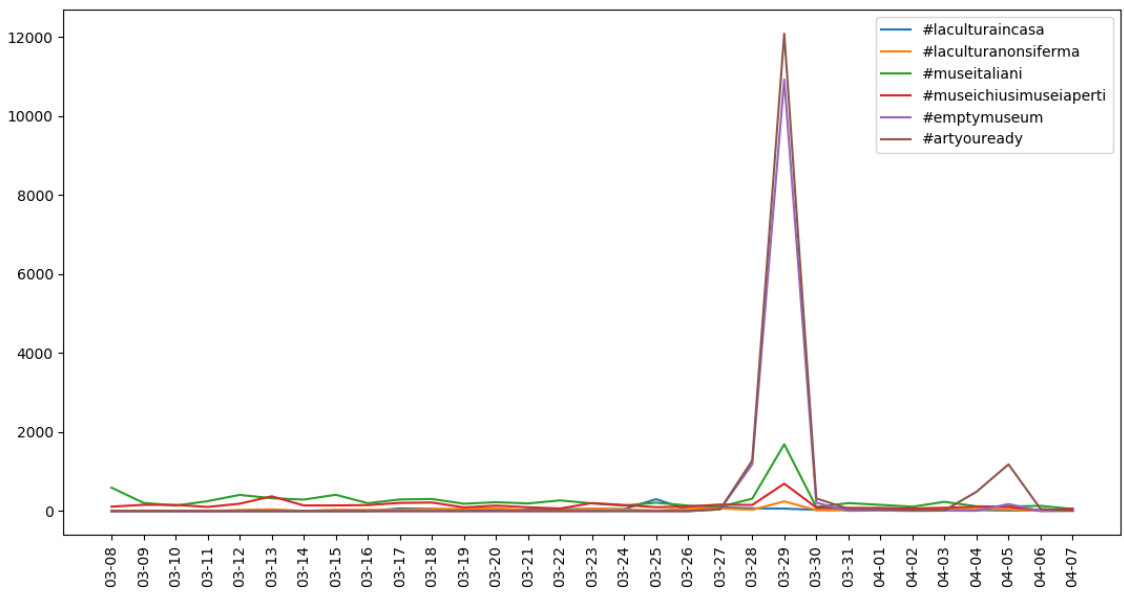

Figure 4 - Retweets' trend per #hashtag limited to the month of March

Figure 5 reports the retweets' trend per *hashtag* regarding the entire month of April and the beginning of May. The interaction growth trend is more or less stable for all the *hashtags*, except for *#ArTyouReady* which was mostly retweeted by users on the 17th, 18th, 19th, 20th, 25th, 26th and 27th of April, and in early May.

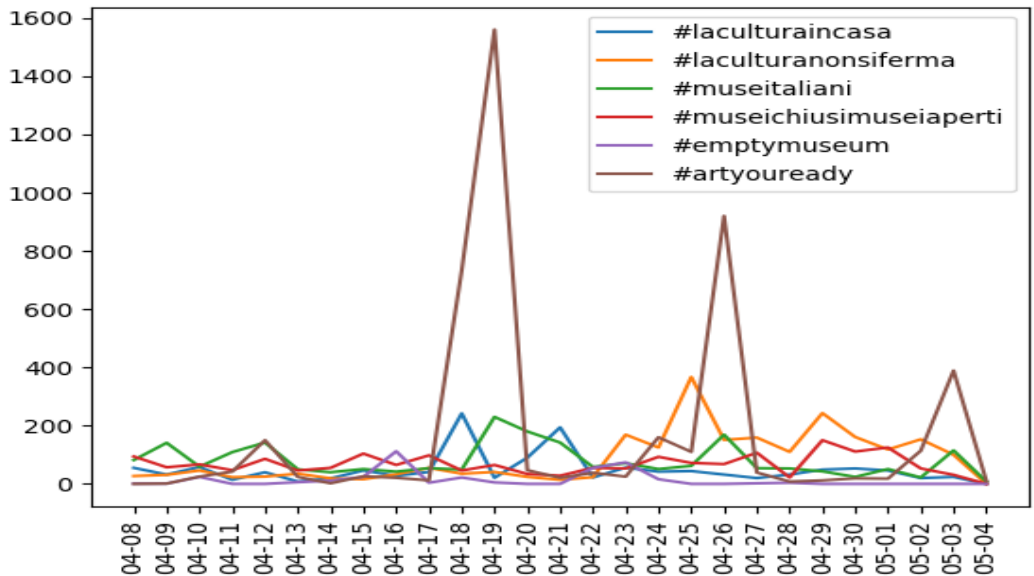

Figure 5 - Retweets' trend per #hashtag limited to the entire month of April and the first four days of May.

Table 3 presents the amount of unique users, namely the number of users who tweet using one or more of the selected #*hashtags*. Data reveal widespread use of #*ArTyouReady*, followed by #*emptymuseum*. However, although these two #hashtags are part of the same initiative and they have been launched by the same cultural institution on the same date, it is worth noticing that #*emptymuseum* was used by about half of those who created tweets containing #*ArTyouReady*. This difference may be due to the fact that #*ArTyouReady* achieved greater fame and has been, for this reason, associated with more content, independently of #*emptymuseum.*

Table 3 - Unique users for each of the six #hashtags considered in the time frame 03/08 - 05/04

| #Hashtag | Unique Users |
|---|---|
| #artyouready | 1456 |
| #emptymuseum | 702 |
| #laculturanonsiferma | 414 |
| #museichiusimuseiaperti | 343 |
| #museitaliani | 282 |
| #laculturaincasa | 215 |
| **Total** | **3412** |
| **Average per #hashtag** | **733.42** |

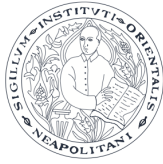

Università degli studi di Napoli
"L'Orientale"

**The institutions involved.** Following the automatic extraction of all users per *#hashtag*, we distinguish two types of categories: 1) ordinary users and 2) cultural institutions. The list of cultural institutions has been manually created and includes 389 institutions, which use at least one of the six *#hashtags* in their tweets. Table 4 shows the number of institutions that singularly used the six *#hashtags*.

Among the identified cultural institutions, we can distinguish 16 classes, grouped according to their type, e.g, archaeological park, library. The Museum class proves to have the largest use of the selected *#hashtags* (Table 4).

Table 4 - Number of institutions per typology that used the single #hashtags

| #Hashtag<br><br>Cultural Institution | #Artyouready | #Emptymuseum | #laculturaincasa | #laculturanonsiferma | #museiitaliani | #museichiusimuseiaperti | Total |
|---|---|---|---|---|---|---|---|
| **Museum** | 10 | 50 | 4 | 77 | 44 | 93 | **278** |
| **Gallery** | 3 | 6 | N/A | 5 | 3 | 5 | **22** |
| **Library** | 1 | 3 | 5 | 4 | 2 | 3 | **18** |
| **Royal Palace** | N/A | 4 | N/A | N/A | 3 | 5 | **12** |
| **Archaeol. Park** | N/A | 3 | N/A | 2 | 3 | 2 | **10** |
| **Public Institution** | 1 | 3 | N/A | 3 | N/A | 2 | **9** |
| **Archive** | N/A | 2 | N/A | 5 | 1 | N/A | **8** |
| **Association** | N/A | 1 | N/A | 4 | 2 | 1 | **8** |
| **Theatre** | 1 | N/A | 4 | 3 | N/A | N/A | **8** |
| **Museum Centre** | N/A | 2 | N/A | 2 | N/A | 3 | **7** |
| **Castle** | 1 | N/A | N/A | 1 | N/A | N/A | **3** |
| **Château** | N/A | 1 | N/A | N/A | 1 | 1 | **3** |
| **Picture Gallery** | N/A | 1 | N/A | 1 | N/A | N/A | **2** |
| **Foundation** | N/A | 1 | N/A | N/A | N/A | 1 | **1** |
| **Total** | | | | | | | **389** |

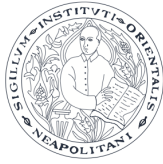

In [Figures 6-10](#), the diagrams show in detail how cultural institutions have used the *#hashtags* during Phase 1. MiBACT resulted to be the most active user of some of these *#hashtags* (i.e. *#ArTyouReady* and *#museitaliani*). Among the local institutions using *#museichiusimuseiaperti* to launch their initiatives, the 'Museo Tattile di Varese' is particularly important as the creator institution of this *#hashtag* and as a small size museum that took advantage of the new channel of communication by broadly using tweets.

Among the institutions that used the analysed *#hashtags*, the presence of well-known cultural attractors is reduced. This reduction might be due to the fact that many of those attractors already exploit specific *#hashtags*, based on their wide-spread identifying names to remark that their identity, so that they show a small use of other *#hashtags* during the pandemic. This is the case of the Archaeological Museum of Naples (MANN) for tweets marked by the *#hashtag #emptymuseum* ([Figure 7](#)).

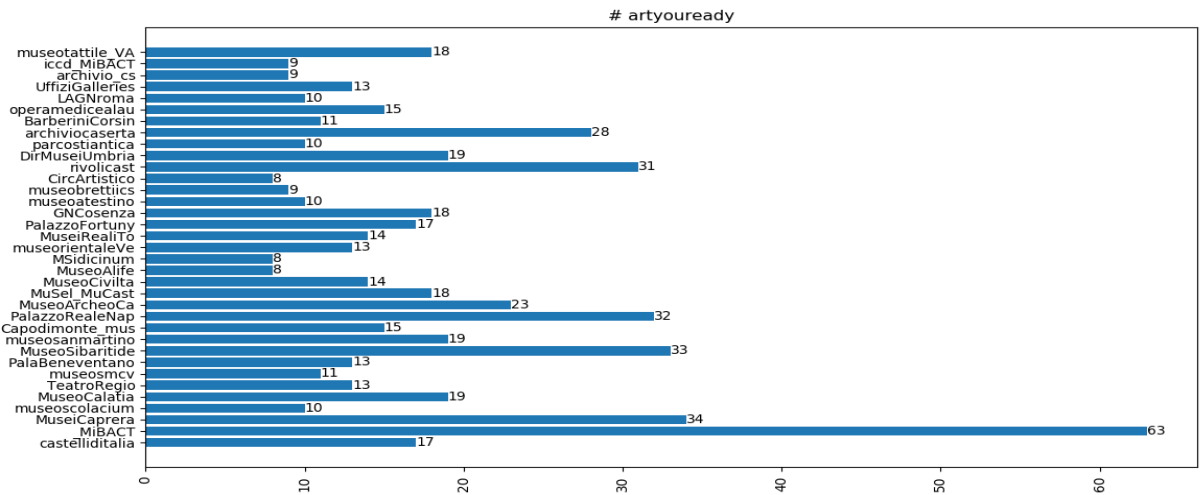

Figure 6 - Names of the cultural institutions using the #ArTyouReady #hashtag

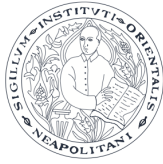

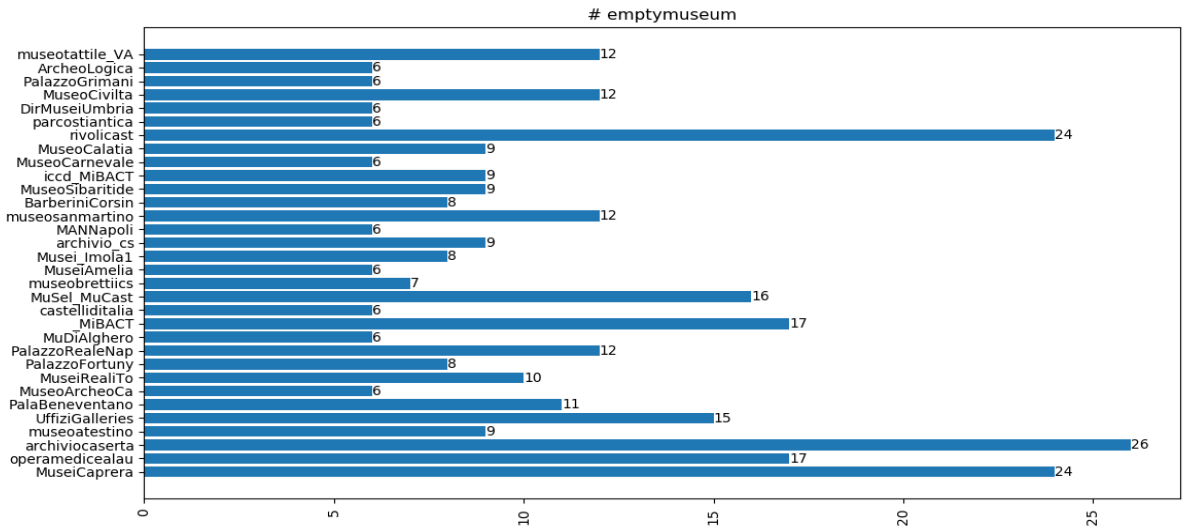

Figure 7 - Names of the cultural institutions using the #emptymuseum #hashtag

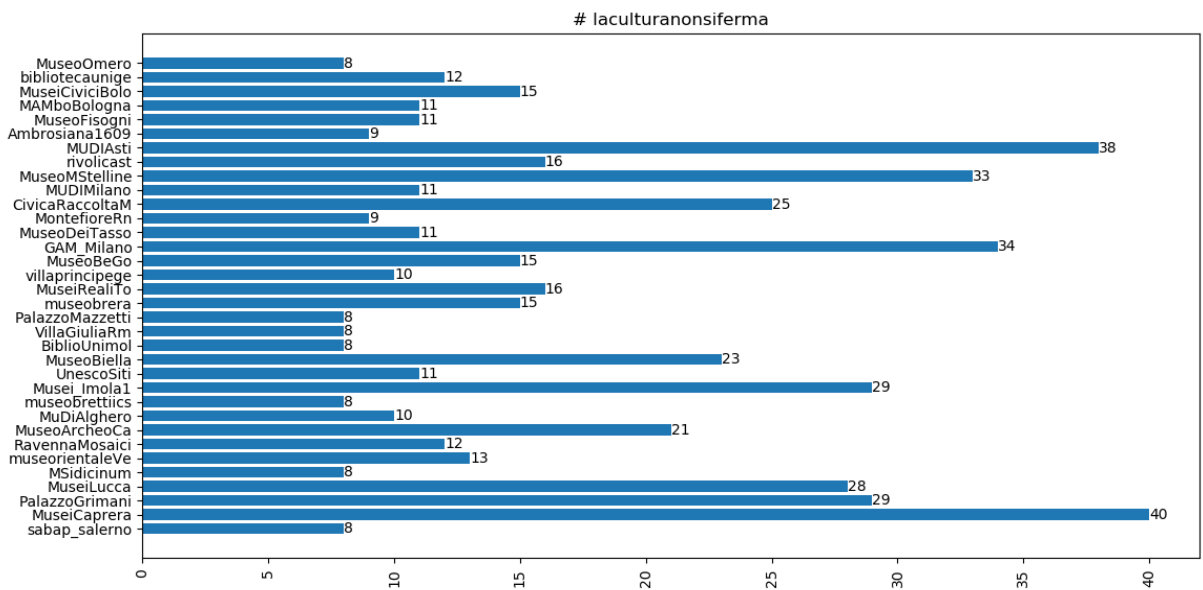

Figure 8 - Names of the cultural institutions using the #laculturanonsiferma #hashtag

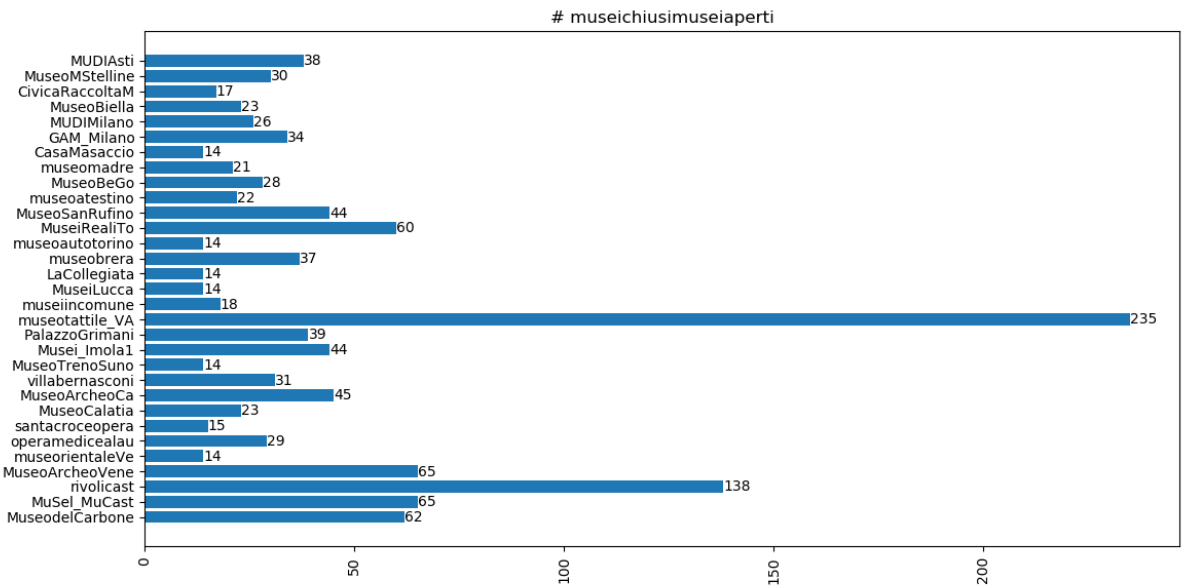

Figure 9 - Names of the cultural institutions using the #museichiusimuseiaperti #hashtag

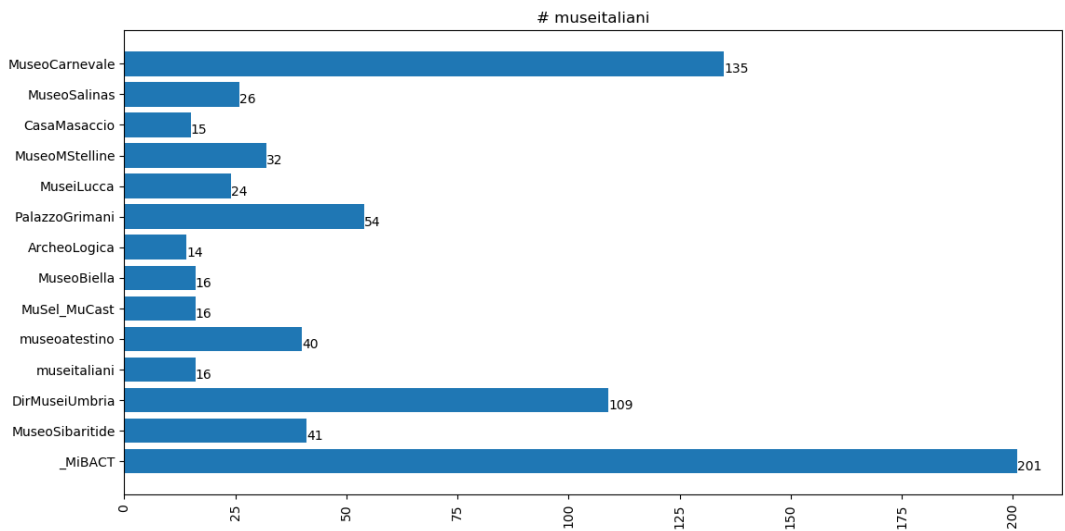

Figure 10 - Names of the cultural institutions using the #museitaliani #hashtag

In Figure 11, we present the data related to users' geographical location. This shows an important concentration of activities in the North and in the Centre of Italy, where the most participating initiatives related to the selectioned *#hashtags* are found. In the South of Italy and in the Islands some activities are registered too, but they cannot be considered particularly relevant, although quite numerous, except for those national resonance institutions (i. e. The Archaeological Museum of Naples - MANN). Three Italian regions, namely Abruzzo, Molise and Basilicata, do not present initiatives related to the selected *#hashtags*.

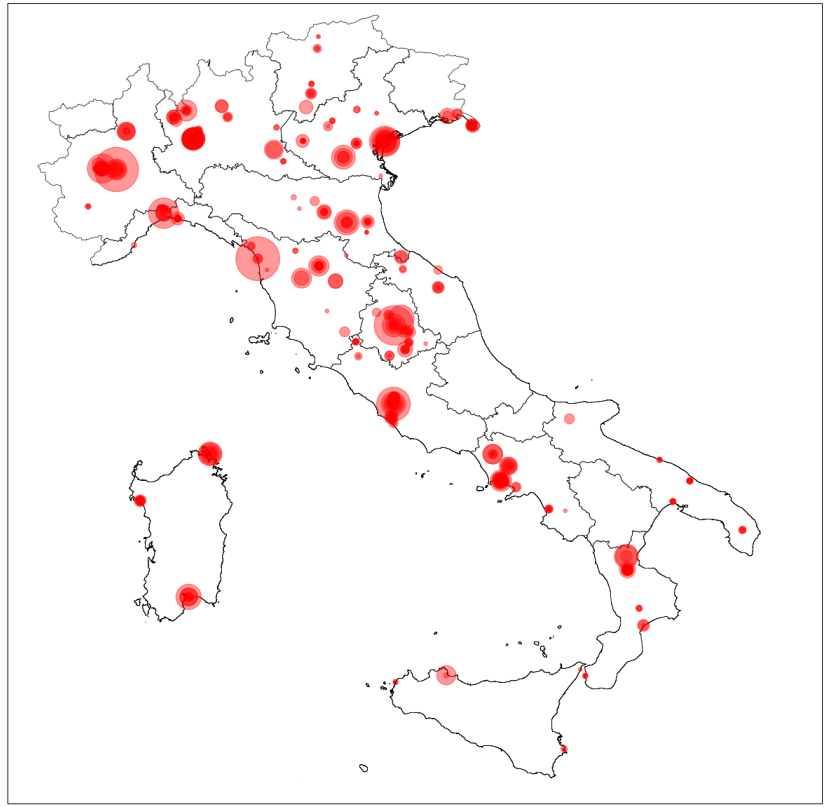

Figure 11- Geographical distribution of the institutions using at least one of the six #hashtags

**Content of #hashtags and tweets.** Regarding to the *#hashtags*' content, it is important to point out that, both for the first and the second dataset, data have been collected without any distinction on the languages used in the related tweets. Thus, a first analysis investigates which languages have been most used as an absolute value compared to all the tweets examined and in relation to the individual *#hashtag*. Obviously, since the initiatives promoted by Italian institutions have been   addressed to a local audience by using aggregators in Italian, the most represented language in a wide number of tweets, as shown in <u>Figure 12</u>. Other languages used are English, French, Catalan, Romanian and Spanish.

In <u>Figure 13</u> the number of tweets related to the *#hashtags* and the languages for each of them are reported. There is a significance presence of the labels *#emptymuseum* and *#ArTyouReady* among the tweets created by English-speaking users.

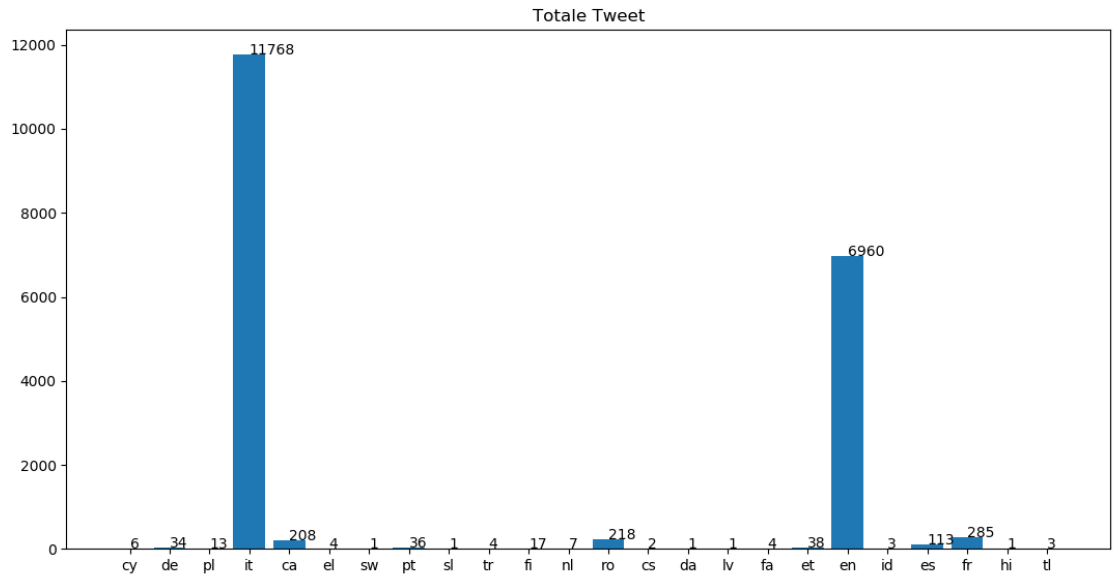

Figure 12 - Number of tweets related to the languages used in the entire dataset

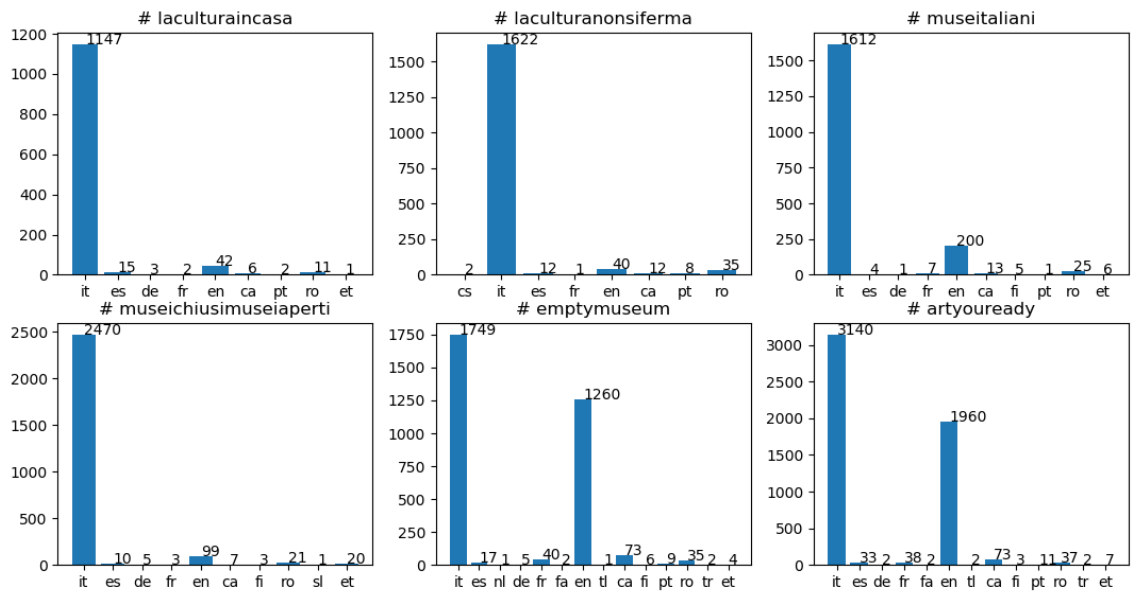

Figure 13 - Number of tweets related to the languages used in each of the six #hashtags

Regarding to the extension of the tweet text the *#hashtags* are associated to, Table 5 reports the tweets' average length for each of the six *#hashtags*. The average has been obtained by counting characters - according to Twitter guidelines that let users to write short texts based on a limited number of characters - not words. It should be noticed that, on average, users have used all the characters available, except for the below-average *#hashtags* *#emptymuseum* and *#ArTyouReady*. This difference can be explained by taking into account the fact that the initiatives related to those *#hashtags* ask users to share images without inserting a descriptive text.

Table 5 - Average length of tweets per #hashtags obtained on the amount of the characters used

| #Hashtag | Tweet average lenght |
|---|---|
| #laculturanonsiferma | 266.58 |
| #laculturaincasa | 263.87 |
| #museitaliani | 258.29 |
| #museichiusimuseiaperti | 249.49 |
| #emptymuseum | 199.67 |
| #artyouready | 197.38 |
| **Total average** | **230.86** |

A first text analysis shows the frequency of words used in tweets tagged with at least one of the selected *#hashtags*. For each of these *#hashtags* a representative wordcloud, showing word occurrences in tweets, has been created (Figure 14).

**#ArTyouReady**

**#emptymuseum**

**#laculturaincasa**

**#laculturanonsiferma**

**#museichiusimuseiaperti**

**#museitaliani**

Figura 14 - Wordcloud for each of the six #hashtags

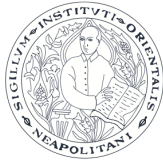

Università degli studi di Napoli
"L'Orientale"

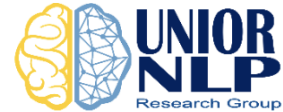

# Conclusion

The report presents the initial results of a survey on the use and diffusion of *#hashtags* by Italian cultural institutions  during the so-called Phase 1 of the lockdown period caused by the COVID-19 epidemic in 2020. The quantitative analysis demonstrated interesting data in terms of both communication from institutions to the audience and users interaction.

The six *#hashtags* that have been selected for the research made it possible 1) to reconstruct most of the initiatives that institutions have nationally and internationally launched through the social network Twitter; 2) to monitor their success throughout the months of March, April and the early May; 3) to identify among the users those who responded to the initiatives the most active institutions on social networks and, finally, 4) to be classify the institutions by thematic areas according to their morphology.

We intend to implement this study in the future, by conducting a linguistic analysis of the content conveyed through the used of such *#hashtags*. We also want to analyze the level of engagement by ordinary users with these initiatives, in terms of *sentiment* and *sensitivity*.